\documentclass[10pt,twocolumn,letterpaper]{article}

\usepackage{iccv}      % To produce the REVIEW version
\usepackage[accsupp]{axessibility}  % Improves PDF readability for those with disabilities.

\usepackage{times}
\usepackage{epsfig}
\usepackage{graphicx}
\usepackage{amsmath}
\usepackage{amssymb}
\usepackage{textcomp} 
\usepackage{gensymb} 
\usepackage{wrapfig}
\usepackage{multirow}
\usepackage{colortbl}
\usepackage{caption}
\usepackage{array}
\usepackage{cuted}
\usepackage{capt-of}

\usepackage{caption}
\usepackage{booktabs}
\usepackage{tikz} 
\usepackage{adjustbox}

\definecolor{iccvblue}{rgb}{0.21,0.49,0.74}
\usepackage[pagebackref,breaklinks,colorlinks,allcolors=iccvblue]{hyperref}

\def\paperID{00003} % *** Enter the Paper ID here
\def\confName{ICCV}
\def\confYear{2025}

\begin{document}

%%%%%%%%% TITLE 
% \title{Window trip}
% \title{Slide }

% \title{FAT-NeRF : Fast And Tileable NeRF for Scaling 3D Neural Fields to Large-Size Remote Sensing Imagery}

\title{Tile and Slide : A New Framework for Scaling NeRF from Local to Global 3D Earth Observation}

% \title{3D tile and Slide: A little slide for SAT-NGP, a big slide for Scaling-Up 3D Earth-Observation reconstruction}

% \title{3D tile and Slide: A little step for SAT-NGP, a big step for Scaling-Up 3D Earth-Observation reconstruction}
% \title{3D tile and Slide: ``One small step for 3D Earth-Observation, one big step for Scaling-Up overhead NeRF''}

% \title{3D tile and Slide: One small step for Scaling-Up overhead NeRF, one big step for 3D Earth-Observation reconstruction"}

\author{Camille Billouard$^{1,2}$\\
% CNES, Univ Gustave Eiffel, ENSG, IGN, LASTIG, F-94160 Saint-Mandé, France\\
{\tt\small camillebillouard314@gmail.com}
\and
Dawa Derksen$^1$\\
% CNES\\
{\tt\small dawa.derksen@cnes.fr}
\and
Alexandre Constantin$^1$\\
% CNES\\
{\tt\small alexandre.constantin@cnes.fr}
\and
Bruno Vallet$^2$\\
% Univ Gustave Eiffel, ENSG, IGN, LASTIG, F-94160 Saint-Mandé, France\\
{\tt\small bruno.vallet@ign.fr}
\and
$^1$ CNES, Toulouse, France \\
$^2$ Univ Gustave Eiffel, ENSG, IGN, LASTIG, F-94160 Saint-Mandé, France
}

\maketitle

\begin{strip}
    \captionsetup{skip=10pt}
    \centering
    \includegraphics[width=0.8\textwidth]{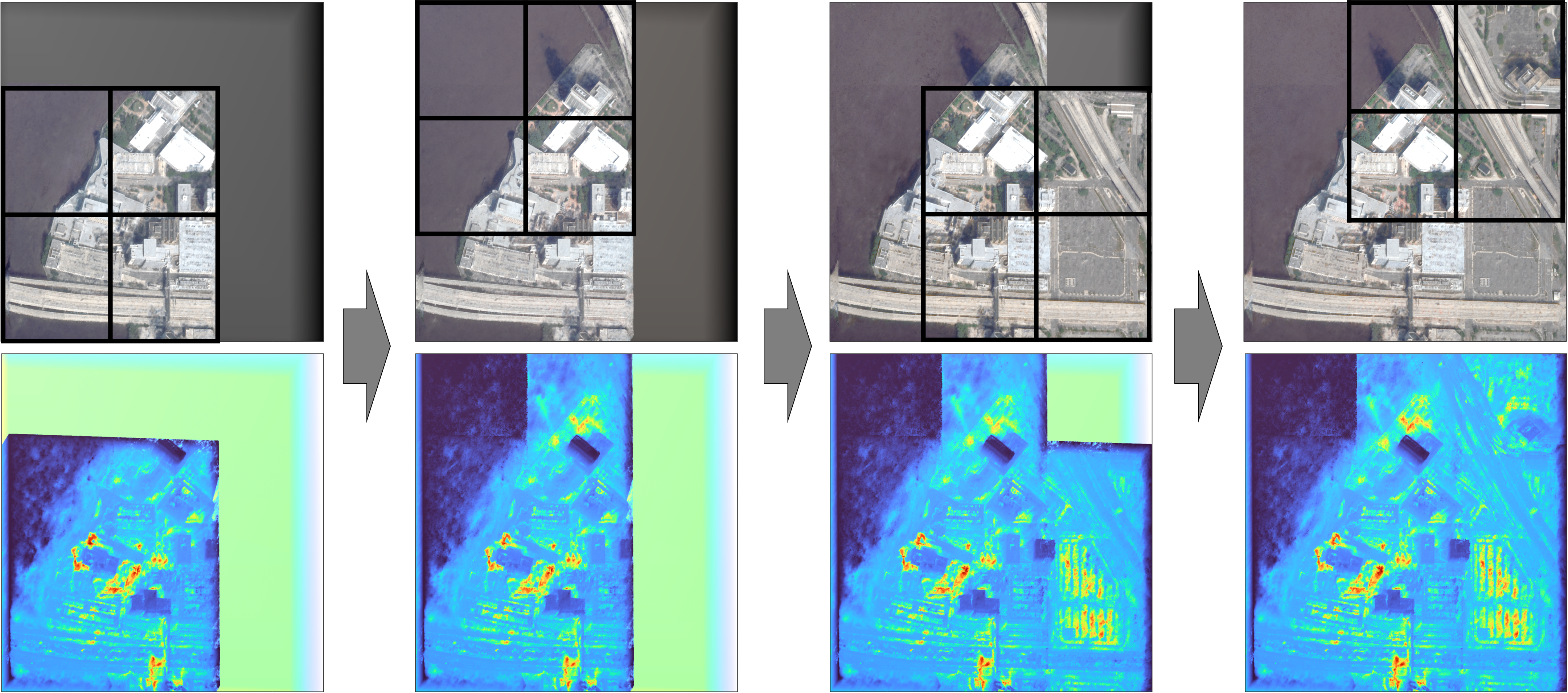}

    \captionof{figure}{We propose a new framework, to scale up NeRFs without scaling up hardware requirements by an out-of-core strategy. Top row represents: Novel View Synthesis (NVS). Second row: the corresponding depth map with highest part in dark blue and lowest in red. The \(2\times 2\) window shows the dynamic adjustment to focus on unexplored regions while maintaining persistent information.}
    \label{fig:main_sans_schéma}
\end{strip}

\begin{abstract}

Neural Radiance Fields (NeRF) have recently emerged as a paradigm for 3D reconstruction from multiview satellite imagery. However, state-of-the-art NeRF methods are typically constrained to small scenes due to the memory footprint during training, which we study in this paper. Previous work on large-scale NeRFs palliate this by dividing the scene into NeRFs. This paper introduces Snake-NeRF, a framework that scales to large scenes. Our out-of-core method eliminates the need to load all images and networks simultaneously, and operates on a single device. We achieve this by dividing the region of interest into NeRFs that 3D tile without overlap. Importantly, we crop the images with overlap to ensure each NeRFs is trained with all the necessary pixels. We introduce a novel \(2\times 2\) 3D tile progression strategy and segmented sampler, which together prevent 3D reconstruction errors along the tile edges. Our experiments conclude that large satellite images can effectively be processed with linear time complexity, on a single GPU, and without compromise in quality. 
% The code is available at \href{https://github.com/Ellimac0/SAT-NGP.git}{our project page}.

\end{abstract}
%%%%%%%%% BODY TEXT
\section{Introduction} \label{sec:intro}

Mapping the Earth in 3D has been a longstanding goal in various communities, spanning from Computer Vision to Remote Sensing~\cite{tuia2024artificial, zhao2022overview}.

In this domain, the Neural Radiance Field (NeRF,~\cite{mildenhall2020nerf}) has emerged as a versatile paradigm for novel view synthesis and 3D reconstruction and demonstrates state-of-the-art performance on street-level imagery ``in the wild''~\cite{chen2022hallucinated, kuang2022neroic, martin2021nerf, zhang2021ners}. More widely, the success of Neural Fields has triggered a large number of studies taking the concept far beyond handheld cameras~\cite{gao2022nerf}. NeRFs are particularly interesting for overhead imagery due to their ability to handle various data sources (\textit{e.g.} LiDAR and optical~\cite{han2025calisa}) within a common 3D representation.

In the context of Remote Sensing, NeRF-based methods promise to handle the challenges of multi-date imagery, including varying lighting conditions~\cite{derksen2021shadow}, transient objects~\cite{mari2022satnerf}, radiometric shifts~\cite{mari2023multi}, and seasonal effects~\cite{gableman2023incorporating}.

Nonetheless, to be effective in operational settings, extensive areas spanning dozens to hundreds of square kilometers need to be mapped in 3D. This raises the question of how to properly scale NeRFs to large areas. Following~\cite{cresson2018large, derksen2019scaling, lassalle2015scalable} we consider that an algorithm is \textit{scalable} under 3 conditions.

\begin{enumerate}
    \item The algorithm should run in linear time (or faster) with respect to the input size : $\mathrm{time} = \mathcal{O}(N)$
    \item The runtime memory should never exceed a fixed memory budget : $\mathrm{memory} = \mathcal{O}(1)$
    \item The tiled algorithm $A_s$ and the reference (untiled) algorithm $A$ should produce similar quality, if not identical results on any input $x$ : $A_{s}(x) \approx A(x)$
\end{enumerate}

The definition of scalability concerns the ability to handle increasing input sizes efficiently. In our case, the input size is the total number of pixels $N = N_{\mathrm{im}}\times A_{\mathrm{ROI}} \times \mathrm{SR}^2$. This paper focuses on scaling to a larger surface area $A_{\mathrm{ROI}}$ while considering as constant the number of images $N_{\mathrm{im}}$, and the Spatial Resolution $\mathrm{SR}$ (in $pix/m$).

Previous research works show that NeRFs do not trivially scale to large scenes~\cite{li2024nerf, tancik2022block, turki2022mega, xu2023grid, zhang2024supernerf}.
Intuitively, larger scenes contain more information, and therefore should require either larger NeRFs, or multiple NeRFs. Considering that the memory footprint during training is linearly proportional to NeRF size, the first solution is impossible (condition 2).
This motivates the need for a principled approach to seamlessly train multiple NeRFs on a large scene with a limited GPU memory budget.

This paper presents a novel approach for scaling NeRFs to large images, specifically adapted to overhead imagery (satellite, airborne, UAV, ...). Our contributions can be listed as follows: 
\begin{itemize}
    \item A tiling strategy based on a square grid according to east/north (UTM coordinates), adapted to large-scale Earth Observation scenes with an extremely low Z variation ($\approx$200m) compared to the other axes ($\approx$10km).
    
    \item A \(2\times 2\) sliding window mechanism to ensure each ray is used at least once in a training batch with all the necessary NeRFs in memory.

    \item A segmented ray sampler that samples per-bounding box, rather than on the entire ray, to guarantee seamless rendering along rays that intersect multiple bounding boxes.
    
    \item Experimental results on a benchmark dataset of Very High Spatial Resolution (VHRS) multi-view satellite images~\cite{lesaux2019data} that highlights:
    \begin{itemize}
        \item The ability to effectively train a set of non-overlapping NeRFs on a single GPU without introducing 3D reconstruction errors along the edges of their domains.
        \item The importance of our design choices, particularly the use of multi-resolution hash tables to encode the geometry of the scene combined with a global color network as in NeRF-XL~\cite{li2024nerf}.
    \end{itemize}

\end{itemize}

\section{Related works} \label{sec:related}

3D models of the surface are useful for many downstream Earth Observations applications, including land cover classification, flood simulation, image orthorectification~\cite{Kuckreja_2024_CVPR}, Digital Twins for disaster response ~\cite{Yu31122023}, environment impact analysis~\cite{Luo06102024}.

\subsection{Neural fields for satellite imagery} \label{sec:related_sat}

In this context, NeRFs have proven to be an elegant solution to handling many of the challenges of multi-date satellite imagery. 

Shadow-NeRF~\cite{derksen2021shadow} tackled the issue of varying lighting conditions, while Sat-NeRF~\cite{mari2022satnerf} improved the camera model and introduced the notion of transient uncertainty. Notable contributions to the field include: EO-NeRF~\cite{mari2023multi} for geometric shadow mapping, Season-NeRF~\cite{gableman2023incorporating} for seasonal effects, BRDF-NeRF~\cite{zhang2024brdf} which incorporates a physics-based Bidirectional Reflectance Distribution Function (BRDF), SparseSat-NeRF~\cite{zhang2023sparsesat} in sparse views context, Sat-mesh~\cite{qu2023sat} for extracting meshes. Recent works~\cite{pic2024pseudo, sprintson2024fusionrf} propose to improve pan-sharpening, or to extract semantic classes~\cite{wagner2025semantic}. Another topic of study is the application of neural fields to Synthetic Aperture Radar~\cite{ehret2024radar}. Across altitudes, NeRFs have been used to extract the 3D geometry of space objects (e.g. satellites, debris) in orbit~\cite{mergy2021vision, heintz2023spacecraft}, on asteroids ~\cite{chen2024asteroid} and on the Moon ~\cite{adams110design}. 
% barbier2025multi 

Several works have taken a key step towards scaling, by accelerating the training time from a dozen hours to a few minutes per scene. This has been achieved using Tensorial Decompositions (Sat-TensoRF)~\cite{zhang2023fast} Neural Graphics Primitives (SAT-NGP)~\cite{billouard2024sat}, and more recently Gaussian Splatting (EOGS)~\cite{aira2025gaussian}. 

While Gaussian Splatting~\cite{kerbl20233d} has shown promise in 3D reconstruction in satellite imagery~\cite{aira2025gaussian}, its application to large-scale imagery presents significant challenges. Specifically, the approximation of satellite sensors as affine cameras, which works well for smaller images, introduces precision errors at larger scales. This leads to inaccuracies in mapping pixel locations to ground coordinates, which is crucial for accurate ray casting. This limitation is critical in the context of large-scale satellite imagery, where high precision is essential for accurate 3D reconstruction.

However, none of these works truly scale to large areas, because they do not respect the 3 conditions given in Section~\ref{sec:intro}. The reason is that for large areas the volume of training rays exceeds the memory budget.

\subsection{Large-scale NeRF} \label{sec:related_scaling}

Much of the research into scaling NeRFs is based on the idea of dividing the scene into multiple NeRFs~\cite{tao2024silvr, turki2023suds, ScaNeRF, Scalable_neural_indoor, 10843328, zhenxing2022switch}. This idea first emerged in Kilo-NeRF~\cite{reiser2021kilonerf}, who proposed to distill a pretrained NeRF into a large number of NeRFs following a 3D voxel grid. This is done in an effort to reduce inference cost and enable real-time rendering. Alone, this idea does not scale to large areas, as it requires a pre-trained ``reference/teacher'' NeRF to distill the color and density outputs directly into the ``student'' NeRFs (without ray rendering). Nonetheless, the idea of sub-division of the scene into smaller, non-overlapping NeRFs is fundamental to our scaling approach. Kilo-NeRF showed that improved rendering speed and identical quality could be achieved with a regular subdivision as long as each NeRFs receives all the information required to learn its 3D domain.

Two relevant steps towards scaling NeRFs were Block-NeRF and Mega-NeRF~\cite{tancik2022block, turki2022mega}. The authors addressed the issue of scaling NeRF using a set of centroids, which equates to dividing the scene into NeRFs with overlap. The rays are rendered according to the NeRFs that they intersect. This is done by combining the inference results weighted by a function of the distance of each sample to the NeRFs centroid. While the edge errors could be handled in this way, we believe that NeRF overlap should be avoided entirely to reduce redundant storage of information and optimize rendering time.

Grid-guided NeRF~\cite{xu2023grid} introduced a hybrid between grid-aligned local NeRFs with a multi-resolution 2D feature grid representation (a Tensorial Radiance Field). While this solution scales elegantly to larger areas due to the pyramidal nature of the representation, they did not address the issue of information division in the feature grid, and scaling to arbitrarily large areas which would require feature grids that would no longer fit into memory. 

Since then, NeRF-XL~\cite{li2024nerf} proposed massively parallel training by dispatching each NeRFs on its own GPU. This work is particularly relevant as they entirely avoid overlap in the NeRFs. This is achieved by introducing a segmented rendering equation that aggregates the transmittance and estimated color of each ray segment into a final pixel color. NeRF-XL introduced the idea of sharing the color network across NeRFs, which regularizes appearance across large areas. 

While these approaches share similarities to ours, one difference with Block-NeRF and Mega-NeRF is that we divide the scene into multiple NGP representations~\cite{billouard2024sat, muller2022instant} without overlap. In contrast to NeRF-XL, we propose a single GPU framework that does not modify the original NeRF rendering equation or the loss function~\cite{mildenhall2020nerf}, in order to accommodate more complex loss functions in the future~\cite{behari2024sundial, derksen2021shadow, mari2022satnerf, martin2021nerf}.

\section{Problem formulation} \label{pb_formulation}
Our fundamental goal is to transform large multi-view satellite images into local, efficient volumetric representations that can be used for 3D reconstruction, novel view synthesis, and to extract domain-specific features of interest (albedo, shadows, transient objects, etc.).

% 200Go images 
% 600Go nerf 
% 1.4To pour rayons
To prove the ability to function within a real-world setting, we constrain our framework to the minimal case: a single GPU device.  Under this hard constraint, entire sets of multi-view satellite images cannot be loaded into memory at once. 
%Typical satellite images encompass extensive areas, often spanning dozens of square kilometers.
For a typical scene of interest (10km by 10km at 30cm Ground Sample Distance) covered by 20 satellite images, the image set weights 200GB and we estimate that a NeRF covering this area at image resolution would weight around 600GB.
Furthermore,  the ray origin and direction vector needs to be stored during training for each ray (in float32 for sufficient precision) which would add 1.4TB. 
Such volumes obviously do not fit on a single GPU. In this paper, we propose solutions to address this problem:
\begin{itemize}
    \item A partition of the 3D scene of interest in individual tiles
    \item A cropping mechanism to load exactly the relevant parts of the images during training. This is different from ray batching, which occurs on each individual partition.
    \item An on-the-fly computation of ray coordinates from the RPCs of each image
\end{itemize}

\subsubsection*{Catastrophic forgetting}
\label{sec:catastro}

Tiling NeRFs require special care to avoid catastrophic forgetting.
We perform a simple experiment to illustrate the impact of naïve image cropping on a single NeRF. First, we divide the area into a \(2\times 2\) grid following the XY (East/North) directions. We call $R$ the ray set that covers the whole scene, and $R_i$ a subset of rays that intersects with a sub-division of the scene, with $i\in\{1, \dots, 4\}$, indicating the lower left, upper left, lower right, and upper right quadrants. For this experiment we train the NeRF in four sequential rounds, using samples from each subsets of rays $R_i$, one at a time.

\begin{figure}[ht]
\begin{center}
   \includegraphics[width=1\linewidth]{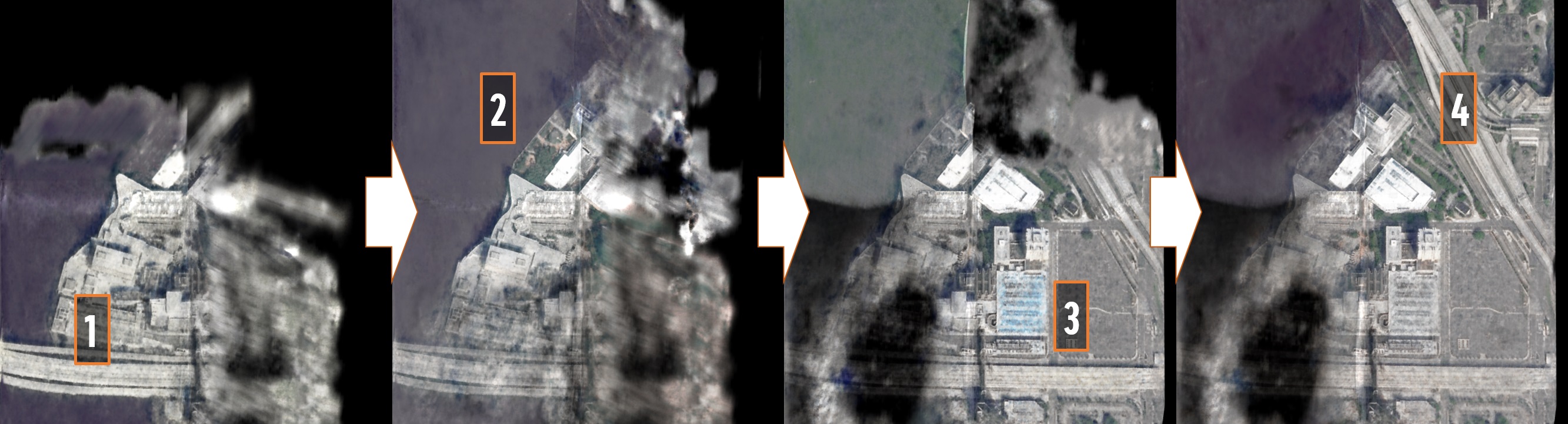}
   
\end{center}
   \caption{Catastrophic forgetting on the scene \textit{JAX 214} from~\cite{lesaux2019data}. We train a NeRF by sampling only from rays in one area at a time. Each time we train a new area, the NeRF progressively forgets the other areas.}
\label{fig:catastro}
\end{figure}

Figure~\ref{fig:catastro} shows that training on a subset of rays $\{R_i, i>1\}$ leads to a loss of information on rays $\{R_j, j<i\}$. This phenomenon is largely known within the Deep Learning community as Catastrophic Forgetting~\cite{kirkpatrick2017overcoming}: when a neural network is re-trained with new data that covers a different region of the input domain than its training data, it may no longer produce coherent results on the previously seen domain.

This experiment shows an example on \textit{JAX 214 scene} where once the image crops of a trained area are unloaded, the NeRF is unable to learn new areas without forgetting the previously learned ones. We propose to overcome this by dividing the scene into non-overlapping NeRFs. This allows us to store \textit{persistent information} in the NeRFs that are already trained. Our approach guarantees by construction that after fully training a NeRFs, the information that it has learned can not be lost.

\subsubsection*{Divide and conquer}

Kilo-NeRF~\cite{reiser2021kilonerf} observes that multiple non-overlapping NeRFs are far more efficient in terms of memory and rendering speed than a single large NeRF~\cite{reiser2021kilonerf}. When the NeRFs are trained with all the necessary information, the rendering quality is equivalent to the single model. Therefore, we seek to propose a straightforward and simple approach to divide the scene into NeRFs, without heavily changing the standard pipeline of NeRF training and inference. 

Second row of Figure~\ref{fig:ideal_naif_ours_9_22} shows that simply tiling the images and reconstructing each area as a separate NeRF incurs errors along the edges of the tiles. Indeed, most 3D reconstruction algorithms rely on a spatial neighborhood of pixels in the images to extract the altitude of the visible surfaces~\cite{beyer2018ames, facciolo2017automatic, leotta2019urban, michel2020new, rupnik2017micmac}. Near the edges of a 3D tile, the spatial neighborhood is incomplete, which leads to errors in the 3D reconstruction.
In many cases, the solution to this problem is to 3D tile the scene with overlapping margins and to stitch the 3D maps along the 3D tile edges. A common strategy is to statistically aggregate the altitude/density/color predictions in overlapping areas~\cite{tancik2022block, turki2022mega}, weighted according to the distance to the 3D tile edge. However, this may incur blurring or other errors in the overlap area~\cite{chen2023scalar, li2024nerf}. 

Instead, we seek to avoid overlap altogether, given that non-overlapping NeRFs are generally faster and more memory-efficient than overlapping NeRFs. As shown in Kilo-NeRF~\cite{reiser2021kilonerf} and NeRF-XL~\cite{li2024nerf}, non-overlapping NeRFs are able to produce fast, high-quality rendering, if they are trained with all of the necessary information.

\section{Proposed method} \label{sec:method}

We aim to define a neural radiance field over a large 3D scene by tiling this scene and defining an individual NeRF on each 3D tile.

First, in Section~\ref{sec:architecture}, we specify the architecture of the NeRFs that we use to describe a radiance field over a 3D tile. 
In Section~\ref{sec:3d_tile_generation}, we present the process of partitioning the 3D scene into non-overlapping tiles.
In Section~\ref{sec:image_crop_from_ground}, we explain the partitioning of images based on their ground coverage.
Next, in Section~\ref{sec:22_slide_window}, we introduce the window displacement process over the entire scene.
Finally, in Section~\ref{sec:seg_sampler}, we detail our segmented sampling mechanism.  
% Finally, in Section~\ref{sec:training}, we discuss the distribution of segments and the training process.

\subsection{NeRFs architecture} \label{sec:architecture}

% local to global 
A NeRF is a network that predicts the color of a scene viewed from a certain point.
This is achieved by generating a 3D ray from each pixel of the rendered scene and sampling points along this 3D ray.
A NeRF receives the coordinates of these samples in the $[0;1]^3$ space mapped to the corresponding ground 3D tile.
We chose the NGP representation~\cite{muller2022instant} in which these coordinates are used to query multi-resolution feature vectors. The feature vectors are provided as input to a local density network that outputs the density estimation at the given coordinate, and a feature vector which we call the density embedding. This embedding, in turn, forms the input of a global color network which is shared across NeRFs. We take inspiration from NeRF-XL~\cite{li2024nerf}, which demonstrates the regularizing effect of a global color network.
% Our ablation study shows the impact of having a local color network and the regularizing effect of the global color network.

Additionally, the NGP features and density networks are trained with individual optimizers and schedulers. The global color network has its own separate optimizer and scheduler. We experimentally find that pursuing learning without resetting the learning rate favors more effective and continuous learning.
During training, we follow the standard procedure of SAT-NGP~\cite{billouard2024sat} and cull samples that fall in empty space using a coarse voxel grid. The voxel density is used to provide the sample density for each ray segment. 

\subsection{3D tiles generation} \label{sec:3d_tile_generation}

The first step is to 3D tile the region of interest (ROI) following a regular square grid~\cite{turki2022mega, xu2023grid} along the east-north UTM coordinates. In our experiments, we consider a simplified scenario where the minimum and maximum Z values are known in advance. These altitude bounds are equal for all bounding boxes.
% Each 3D tile $T_{n,m}$ is an axis aligned square.

\begin{figure}[ht]
\begin{center}
   \includegraphics[width=1\linewidth]{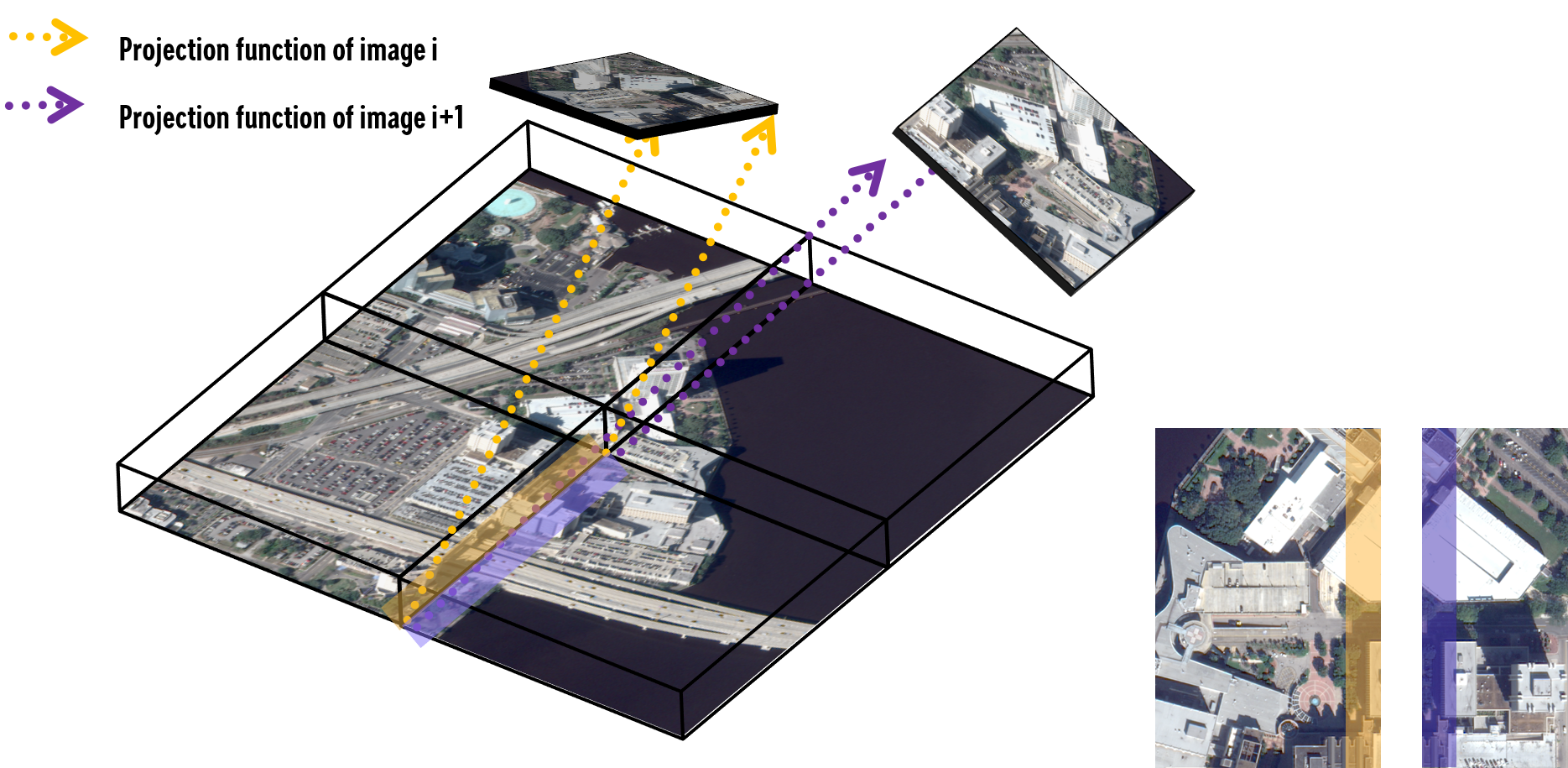}
\end{center}
   \caption{ Individual non-overlapping tiles are defined relative to their specific UTM partitions, ensuring contiguous spatial segmentation. The partitioning of images is based on their ground coverage, taking into account the minimum and maximum altitudes. The highlighted areas in orange and purple indicate regions where the projections of the 3D tiles in the image overlap. 
   % The highlighted areas in orange and purple, corresponding to area view by one of each area, indicate regions of the image with overlap.
   }
\label{fig:3D_partition_merge_tuilage_from_roi}
\end{figure}
% This tiling approach partitions into square tiles, providing a simple and efficient spatial segmentation. Although it may not provide the optimal tiling, its simplicity and uniformity bring consistent spatial divisionin UTM coordinate systems. 

% je vais expliquer qu'on se sers des fonctions de projection ou loc onverse pour faire ça: appuyé par un schéma 

These coordinates will be used to crop the images, and therefore define the ray batches. Moreover, we perform UTM-to-local transformation of ray origins and directions to ensure the samples fall within the $[0;1]^3$ space known to be required for NeRF training~\cite{derksen2021shadow}. Figure~\ref{fig:3D_partition_merge_tuilage_from_roi} illustrates this 3D tiling.

\subsection{Image crop from ground tiles}  \label{sec:image_crop_from_ground}

Thanks to the RPC camera model~\cite{rpc_sar}, we can project points from the 3D space to image coordinates. For each 3D tile and each input image, we create a crop of the images containing the projections of the 8 corners of the 3D tile. To limit the memory footprint, when processing a set of tiles, we will load only the corresponding NeRFs and crops. This approach guarantees that the 3D tile of each loaded NeRF is fully covered by the loaded images (see Figure~\ref{fig:3D_partition_merge_tuilage_from_roi}). This alleviates the need to fully load an entire large image into memory at any moment, something that is commonly overlooked in NeRF pipelines. 

\subsection{Sliding window} \label{sec:22_slide_window}

\begin{wrapfigure}{r}{0.15\textwidth}
  \vspace{-20pt}
  \begin{center}
    \includegraphics[width=0.15\textwidth]{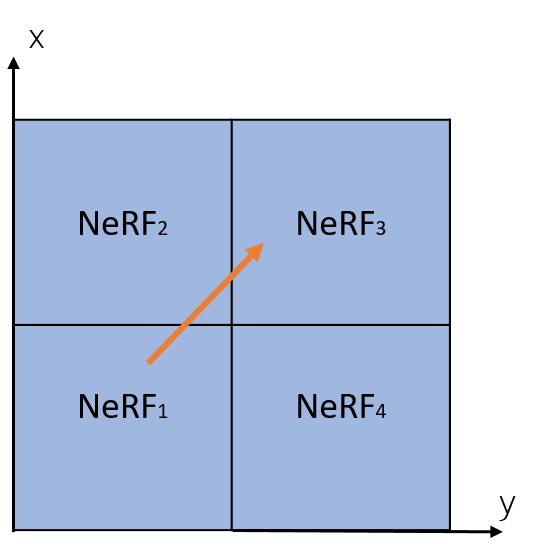}
  \end{center}
  \vspace{-10pt}
  \caption{View from above, orange ray intersecting a maximum of 3 NeRFs.}
  \vspace{-10pt}
  \label{fig:intersect_aabb}
  
\end{wrapfigure}

In the overhead scenario, given the low incidence angle of the satellite, each ray can intersect at most 3 tiles. We call the rays intersecting more than one 3D tile \textit{shared rays} (Figure~\ref{fig:intersect_aabb}). 
Consider the setting where a single NeRF is trained at a time. 
In this scenario, if the shared rays are included as training rays, they will not be able to propagate any gradient onto the neighboring NeRFs that are not loaded in memory. This means that objects outside the bounding box will be explained as ``floaters'' (density estimation errors) inside the bounding box, as shown in Figure~\ref{fig:ideal_naif_ours_9_22}. 
On the other hand, excluding shared rays altogether results in a low information density along the edges of the 3D tile also incurs density estimation errors. This phenomenon occurs systematically along the edges of the overall region of interest (Figure~\ref{fig:ideal_naif_ours_9_22}).
Therefore, without overlap, loading multiple NeRFs into memory becomes necessary at some point during training to properly learn the edges between their domains.

% For this reason we propose to train 4 NeRFs at a time, following a \(2\times 2\) window.

\begin{figure}[ht]
\begin{center}
   \includegraphics[width=0.75\linewidth]{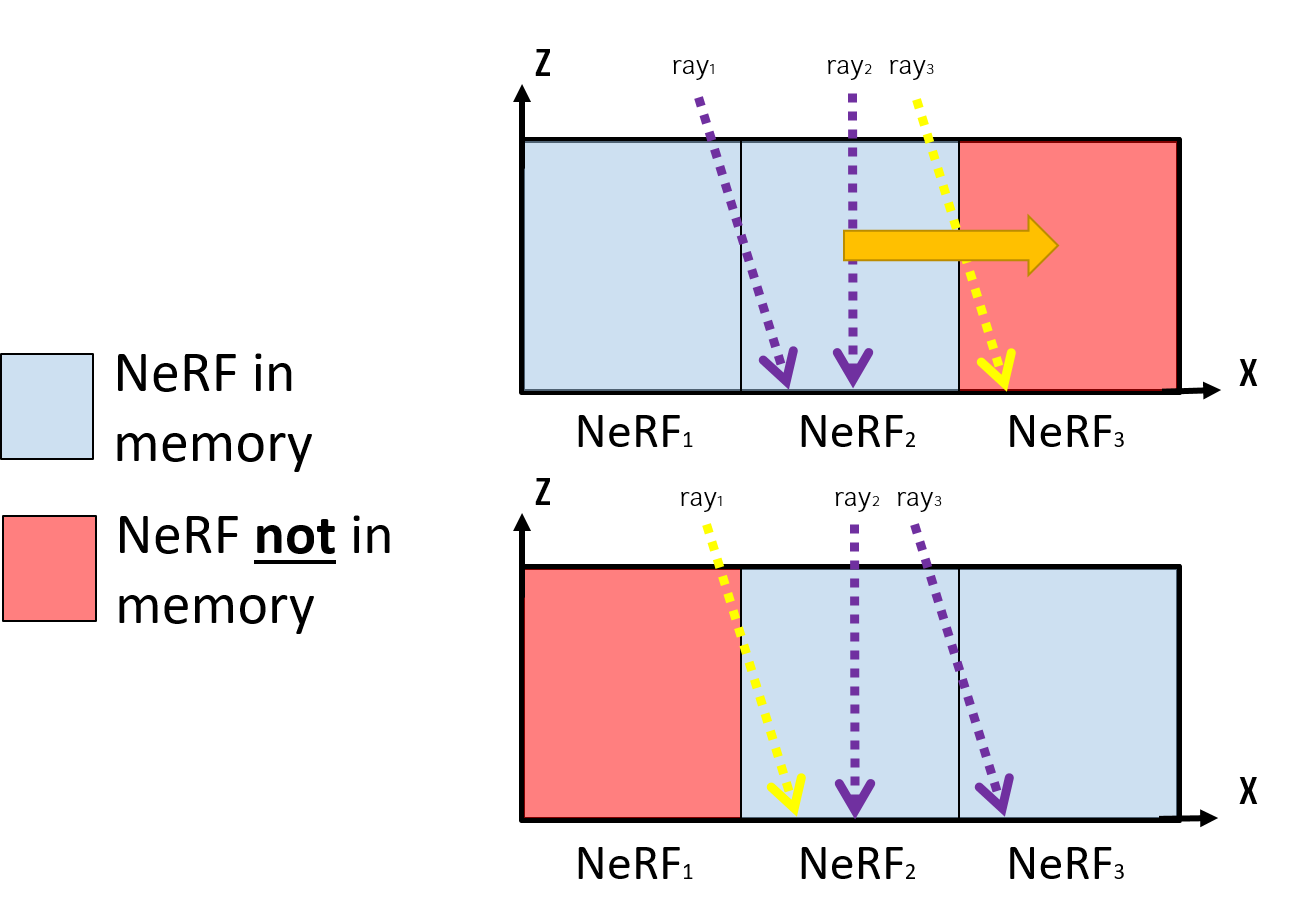}
\end{center}
   \caption{Rays (purple arrows) intersecting only bounding box currently in memory (blue), will be accepted for sampling. Conversely, rays (yellow arrows) intersecting a NeRF that is not currently in memory (red) will not be accepted.}
\label{fig:sliding_window_accepted_rays}
\end{figure}

For this reason and to ensure optimal utilization of GPU memory, regardless of its capacity, we propose a consistent approach of training 4 NeRFs at a time, following a \(2\times 2\) window.
Importantly, we include rays that fall within any of the 4 NeRFs loaded in memory, and exclude rays that intersect with NeRFs not loaded in memory. See Figure~\ref{fig:sliding_window_accepted_rays} for a simple 1D example.

Near the edge of the \(2\times 2\) area, the rays that intersect NeRFs not loaded in memory are excluded. This implies a lower information density in this area.If the area has not yet been learned, this introduces reconstruction errors. However, these excluded rays will be accepted and processed when the sliding window advances over them, thereby correcting any reconstruction errors.
% If the area has not yet been learned, this introduces reconstruction errors which will later be corrected when the sliding window advances. 
If the area has already been learned, we find that the low information density does not cause catastrophic forgetting. Applying the standard NeRF color consistency loss~\cite{mildenhall2020nerf} on rays that intersect only the NeRF in memory is sufficient to preserve learned information.
\begin{wrapfigure}{r}{0.13\textwidth}
  \vspace{-20pt}
  \begin{center}
    \includegraphics[width=0.13\textwidth]{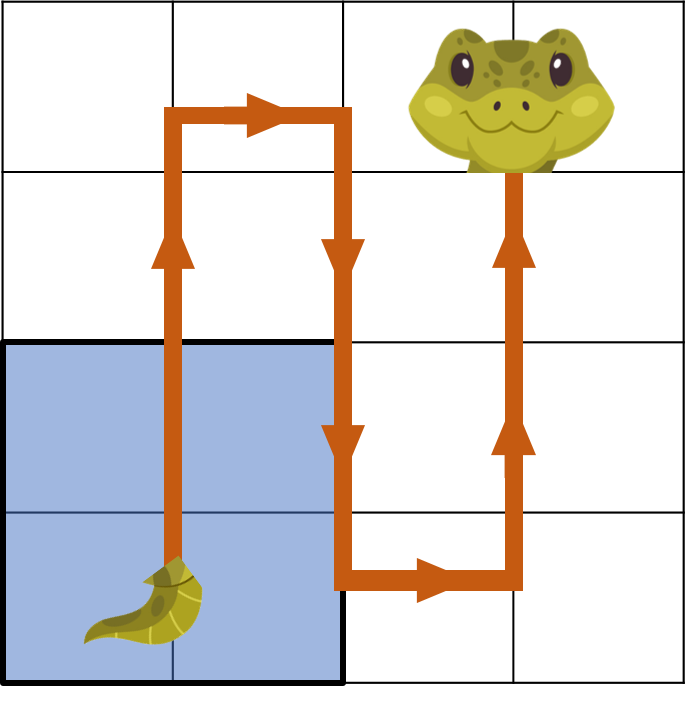} %  fun 

  \end{center}
  \vspace{-5pt}
  \caption{Window sliding path in a \(4\times 4\) grid.}
  \vspace{-10pt}
  \label{fig:window_path_slide}
\end{wrapfigure}
After a fixed number of iterations, we slide the \(2\times 2\) window across the scene to cover the entire area, following the "snake" pattern in Figure~\ref{fig:window_path_slide}. This implies that each NeRF is trained 1, 2 or 4 times respectively for corners, edges, and central tiles. This strategy maintains high spatial locality by reducing the distance between grid areas to be processed and minimizing the need to reload recently processed data. This strategy ensures that only the 2 adjacent NeRFs models are loaded or unloaded into memory.
A key to the success of our method relies on continuing the training rounds without resetting the optimizer. If the rays that the NeRF has already seen are discarded from the training batch, the NeRF suffers from catastrophic forgetting, illustrated in Section~\ref{sec:catastro}. Presenting the NeRF with information it has already received is akin to a recall strategy that is commonly used in Continual Learning~\cite{cai2023clnerf, po2023instant, singh2024c, wang2024scarf, zhang2023nerf}. 

Once the rays that intersect all 4 NeRFs are computed, we randomly select a batch of rays for stochastic gradient descent. We perform the standard NeRF procedure on these rays: sampling, rendering, and propagating the color consistency loss~\cite{mildenhall2020nerf} onto the parameters of the representation.

\subsection{Segmented sampler} \label{sec:seg_sampler}

\begin{figure}[ht]
\begin{center}
   \includegraphics[width=0.6\linewidth]{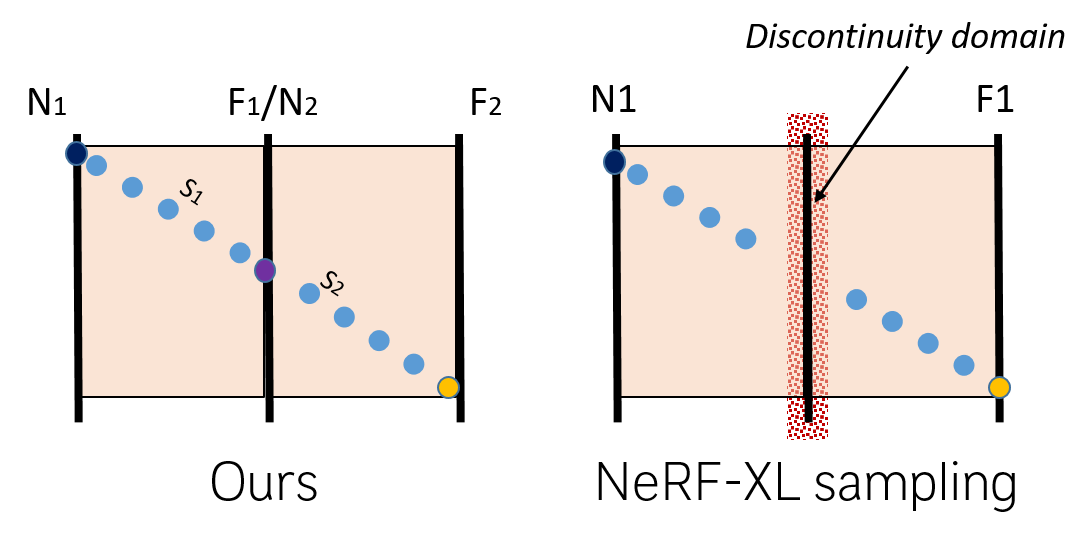}
\end{center}
  \vspace{-5pt}

   \caption{Ray segmentation before sampling (\textbf{left}) ensuring domain continuity, in contrast to a ray that is sampled and then distributed based on its belonging to the NeRF (\textbf{right}). $N_i$ for near $i$, $F_i$ for far $i$ and $S_i$ for segment $i$}
\label{fig:aabb_near_far_seg_sampler_horizon}
\end{figure}

A key component of our algorithm is to sample along an entire ray which is crossing multiple NeRFs. For each NeRF viewed by a ray, we identify near/far (resp. N/F) rays as shown in Figure~\ref{fig:aabb_near_far_seg_sampler_horizon}.

\begin{figure}[ht]
\begin{center}
   \includegraphics[width=0.7\linewidth]{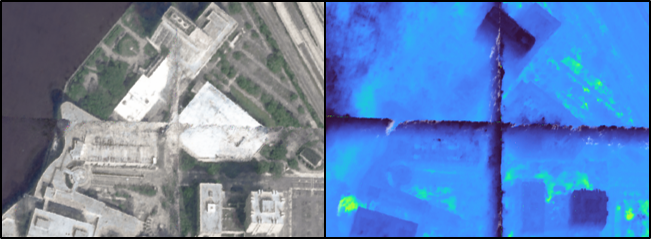}
\end{center}
   \caption{\textbf{Left}: novel view synthesis, \textbf{Right}: depth map. When training with uniform sampling (NeRF-XL,~\cite{li2024nerf}), we observe a thin wall of hallucinated matter along the edges of the tiles.}
\label{fig:nerf_xl_samplingrendering}
\end{figure}

Figure~\ref{fig:aabb_near_far_seg_sampler_horizon} shows the difference between our proposed sampling and NeRF-XL sampling distribution across models. Rather than sampling along the entire ray (and distributing the samples across NeRFs) we propose to sample separately on each ray segment $S_1$ and $S_2$, after identifying near/far points at the edges of the tiles. We find that this avoids false creation of matter near the limit between two segments (see Figure~\ref{fig:nerf_xl_samplingrendering}). This stems from the observation that NeRFs are unable to precisely learn near the borders of a 3D domain if the near/far points do not fall exactly on the limits of this domain. In the Figure~\ref{fig:aabb_near_far_seg_sampler_horizon}, our method on the left guarantees continuity of the domain (purple dot), whereas the method on the right does not ensure this continuity and introduces uncertainty at the boundaries. We believe this phenomenon may be accentuated in our case due to the sparse number of views (10--20) compared to the datasets used in NeRF-XL with typically hundreds of views.

One may observe that the point at the intersection between the ray and the bounding box is sampled twice. First as the far bound of $S_1$ and again as the near bound of $S_2$. Considering that the NeRF rendering (Equation (3)~\cite{mildenhall2020nerf}) takes into account the $\delta x$ between successive samples, the density estimation at $N_2$ will be ignored in the rendering. Thankfully, the bounding box is crossed from both sides, so even with relatively sparse views, the NeRFs at either side are able to correctly learn the density along the edge. 

\subsection{Implementation} \label{sec:implementation}

We release an open-source highly parallelized implementation of our method, leveraging the computational power of modern GPUs to achieve efficient and scalable performance thanks to the following contributions :
%. Our approach includes several key contributions
\begin{itemize}
    \item \textbf{On-the-fly Ray Calculation}: We compute the origin and direction of rays on-the-fly, in a massively parallelized manner with custom CUDA kernel. To the best of our knowledge, other methods in~\ref{sec:related_sat}, pre-calculate all rays in float32 (crucial for maintaining high precision). Our approach avoids storing all rays in memory.

    \item \textbf{Dynamic Segment Distribution}: The distribution of segments is also performed utilizing massively parallelized custom CUDA kernel. In our parallelized approach, computation times are reduced by a factor of 10 compared to sequential and non-parallelized settings, underscoring the efficiency of our method.

    % \item \textbf{New Framework Implementation:} Since our images are not fully loaded into memory, we introduce a novel framework designed to handle batch random processes within a subspace, enhancing the flexibility and efficiency of our method.
\end{itemize}

Our \href{https://github.com/Ellimac0/Snake-NeRF}{implementation} is built upon known frameworks : Nerfstudio~\cite{tancik2023nerfstudio} and NerfAcc~\cite{li2023nerfacc}. By employing custom CUDA kernels, we achieve significant speed-ups and optimizations. The source code and all related data will be available, ensuring reproducibility and further research. The training procedure shares the same loss function and optimization strategy as I-NGP~\cite{muller2022instant}.

\section{Experimental results} \label{sec:results}

\subsection{Dataset} \label{sec:dataset}

To demonstrate the validity of our approach in a controlled setting we base our experiments on the IEEE 2019 Data Fusion Contest~\cite{lesaux2019data}. This dataset contains over 60 scenes with between 10 and 20 pan-sharpened multi-view Worldview-3 images at a 0.3m spatial resolution over Jacksonville and Omaha, USA. 
% The dataset also contains an aerial LiDAR-derived Digital Surface Model (DSM) that provides a ground truth for the 3D reconstruction task on a small part of the scene.  

%For each scene evaluated using our method, the total number of pixels, their footprint, and the footprint of each NeRFs in $km^2$.

% \begin{table}[t]
% \centering
% \begin{tabular}{lccc}
% \toprule
%  & Number of Pixels & Ground Coverage \\
% \midrule
% Scene 214 & $84.910.080$ & Coverage / one NeRF \\
% Scene 068 & $n$ & Coverage / one NeRF \\
% Scene 004 & $n$ & Coverage / one NeRF \\
% Scene 160 & $n$ & Coverage / one NeRF \\
% % Scene xxx & $n$ & Coverage / one NeRF \\
% % Scene xxx & $n$ & Coverage / one NeRF \\
% % Scene xxx & $n$ & Coverage / one NeRF \\
% % Scene xxx & $n$ & Coverage / one NeRF \\
% \bottomrule
% \end{tabular}
% \caption{For each scene, the number of total pixels across images, and their ground coverage in km.}
% \label{tab:scenes_infos}
% \end{table}

\subsection{Experimental setup} \label{sec:setup_expe}

Our goal is to prove the capacity of our approach to effectively scale to large areas following the 3 conditions which we remind here. First, the computation time should be at most linear with respect to the input size, in our case, the area of the region of interest. Second, the memory footprint should never go beyond a fixed limit. Third, the result of the scaled and unscaled algorithms should at least be similar, if not identical. 

The NeRF framework allows us to control the computation time by increasing or decreasing the number of iterations. Suppose each training iteration takes a time $t_{\mathrm{it}}$.
We fix the number of iterations per NeRF to $n_{\mathrm{it}}$, which results in a linear time complexity with respect to the total number of NeRFs per row ($H$) and column ($W$), $N_{\mathrm{nerfs}} = HW$, used over the area of interest. 

$$\mathrm{time} = t_{\mathrm{it}}N_{\mathrm{nerfs}}N_{\mathrm{it}} = t_{\mathrm{it}}HWn_{\mathrm{it}}.$$

While we can select any values for $H$ and $W$, we postulate that the optimal value of $HW$ should increase proportionally (w.r.t.) the area of the region of interest. 

The second condition is theoretically verified due to the fact that we only need to load ray subsets and 4 NeRFs at a time in memory, which can be dimensioned to fit within any GPU memory. 

The third condition is difficult to prove theoretically given the stochastic nature of gradient descent. With different initializations, the NeRFs will not converge to a strictly identical solution as a global NeRF trained on the same data. We therefore seek to verify the third condition empirically by showing that the scaling approach does not introduce any significant reconstruction errors, particularly along the edges of the NeRFs. 
 
% To verify this property we need to compare the results between the scaled and unscaled versions (aka ideal case with unlimited memory) of the algorithm. This cannot be accomplished on a large area given that the unscaled algorithm needs to be executable for the comparison. Therefore we focus this comparison on a small area and verify the equivalency of the scaled and unscaled algorithms on this small area. The unscaled algorithm is equivalent to running a single NeRFs unit on the entire scene.

To quantify the similarity of the scaled and unscaled algorithms, we compare the images Novel View Synthesis (NVS) and depth maps (DSM) of :
\begin{itemize}
    \item The \textbf{reference} algorithm (ideal case with unlimited memory), equivalent to running a single NeRF on the entire scene with all images at same time, % pour bien rappeler que le cata oubli c'est pas ça
    
    \item \textbf{Unscaled} algorithm training the NeRFs one at a time.

    \item Our scaled approach, \textbf{Snake-NeRF} (Section~\ref{sec:method}). 
\end{itemize}

These experiments cannot be accomplished on a large area given that the reference needs to be executable for the comparison. Therefore we focus this comparison on four small areas and verify the equivalency of the scaled and reference algorithms on different urban landscapes. 

\subsection{Evaluation metrics} \label{sec:metrics}

We evaluate the quality of our scaling algorithm in 2 distinct ways. First we evaluate the \textit{absolute quality} of the generated images and DSMs by comparing the reference, unscaled, and scaled versions to the ground truth test images and LiDAR DSM. We consider scaling to be successful if the scaled algorithm reaches the same performance as the reference algorithm.

Second we evaluate the \textit{relative quality} of the scaled and unscaled algorithms, considering the reference algorithm as a reference. We seek to quantify the error introduced by the unscaled algorithm, independently of the errors of the reference.
We perform this evaluation on the Novel View Synthesis task by measuring the Peak Signal to Noise Ratio / Structural SIMilarity (PSNR/SSIM) between images, as well as on the 3D reconstruction task by measuring the Mean Absolute Error (MAE) between the depth maps. 

\definecolor{gold}{rgb}{1.0, 0.75, 0.0}
\definecolor{silver}{rgb}{0.75, 0.75, 0.75}
\definecolor{bronze}{rgb}{0.8, 0.5, 0.2}

\begin{table*}[!ht]
\centering
\caption{Quantitative evaluation of our proposed method across four distinct JAX datasets. The table presents absolute PSNR and SSIM values relative to the ground truth, alongside comparative metrics against a reference algorithm. Optimal results are highlighted in bold. Additional comparative analyses are provided in the supplementary materials.}
\renewcommand{\arraystretch}{1.5} 
\resizebox{\textwidth}{!}{%
\begin{tabular}{l|ccc|ccc|ccc|ccc}
\toprule
 & \multicolumn{3}{c|}{JAX 214} & \multicolumn{3}{c|}{JAX 175} & \multicolumn{3}{c|}{JAX 165} & \multicolumn{3}{c}{JAX 168} \\
\cmidrule(r){2-4} \cmidrule(lr){5-7} \cmidrule(lr){8-10} \cmidrule(l){11-13}
Methods & \begin{tabular}{@{}c@{}}PSNR\\ (abs/rel)\end{tabular} & \begin{tabular}{@{}c@{}}SSIM\\ (abs/rel)\end{tabular} & \begin{tabular}{@{}c@{}}MAE\\ (rel)\end{tabular} &
\begin{tabular}{@{}c@{}}PSNR\\ (abs/rel)\end{tabular} & \begin{tabular}{@{}c@{}}SSIM\\ (abs/rel)\end{tabular} & \begin{tabular}{@{}c@{}}MAE\\ (rel)\end{tabular} &
\begin{tabular}{@{}c@{}}PSNR\\ (abs/rel)\end{tabular} & \begin{tabular}{@{}c@{}}SSIM\\ (abs/rel)\end{tabular} & \begin{tabular}{@{}c@{}}MAE\\ (rel)\end{tabular} &
\begin{tabular}{@{}c@{}}PSNR\\ (abs/rel)\end{tabular} & \begin{tabular}{@{}c@{}}SSIM\\ (abs/rel)\end{tabular} & \begin{tabular}{@{}c@{}}MAE\\ (rel)\end{tabular} \\
\midrule
Reference case                              & 19.10/- & 0.90/- & -                                  & \textbf{19.72}/- & 0.83/- & -                  & 16.21/- & 0.79/- & -         & 19.34/- & \textbf{0.84}/- & - \\
\hline
Unscaled                                    & 15.82/16.22 & 0.72/0.78 & 0.07                            & 14.4/15.53 & 0.59/0.62 & 0.08                 & 15.13/15.97 & 0.61/0.68 & 0.09         & 15.90/16.42 & 0.74/0.79 & 0.06 \\
\hline
Ours: grid \(3\times 3\)   & \textbf{19.30}/21.80 & 0.86/0.93 & \textbf{0.06}       & 19.45/\textbf{26.14} & 0.80/\textbf{0.94} & \textbf{0.03}         & 17.3/23.86 & 0.81/0.94 & 0.03         & 19.20/26.02 & 0.80/0.93 & 0.03 \\
\hline
Ours: grid \(4\times 4\)   & 19.16/\textbf{23.22} & 0.87/\textbf{0.95} & \textbf{0.06}       & 18.96/24.42 & 0.77/0.91 & 0.04         & \textbf{17.9}/\textbf{24.10} & 0.80/\textbf{0.95} & \textbf{0.04}         & \textbf{19.53}/\textbf{26.09} & 0.81/\textbf{0.94} & \textbf{0.02} \\
\bottomrule
\end{tabular}%
}

\label{tab:expe_metrics}
\end{table*}

\subsection{Results and analysis} \label{sec:results_analysis}

\begin{figure}[ht]
\begin{center}
   \includegraphics[width=1\linewidth]{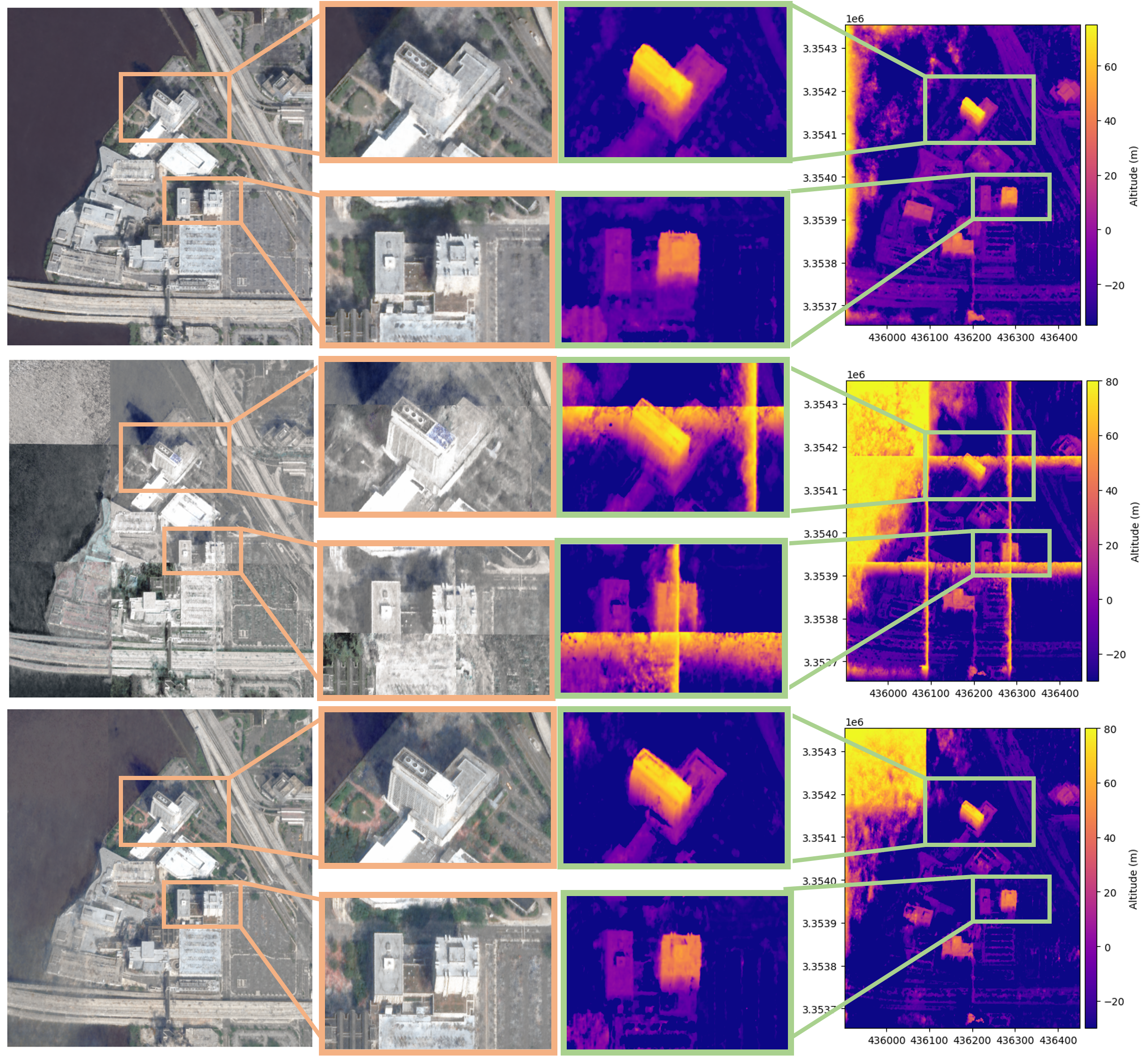}

\end{center}
   \caption{The \textbf{reference} algorithm (\textbf{Top}), utilizes a single model to learn the entire scene. Errors are caused by shadows and transient effects. The \textbf{unscaled} case (\textbf{Middle}) where a single NeRFs is trained at a time, demonstrates significant artefacts along the 3D tile edges. \textbf{Snake-NeRF} (\textbf{Bottom}) shows similar performance to the reference algorithm, while being applicable to large areas.}
\label{fig:ideal_naif_ours_9_22}
\end{figure}
Figure~\ref{fig:ideal_naif_ours_9_22} illustrates the Novel View Synthesis on an unseen test image (\textbf{left}) and the depth map (\textbf{right}) for the three methods we compare. In the top row, we show the reference algorithm, where the entire scene is contained in a single block. We observe that errors occur in areas affected by shadows and transient objects.  

%In the depth map, dark blue regions indicate the highest elements in the scene, while red regions denote the lowest areas, with the latter revealing "holes" due to the exclusion of transient objects and shadows.

The second row displays unscaled algorithm, that involves training each NeRF independently, while including rays that intersect neighboring bounding boxes. This clearly introduces boundary reconstruction errors, most evident in the depth map where density errors appear on the bounding box edges. This occurs because the NeRFs attempt to explain density and color that is outside their bounding box by hallucinating matter within their bounding box, as mentioned in Section~\ref{sec:22_slide_window}.

\begin{figure}[ht]
\begin{center}
   \includegraphics[width=0.7\linewidth]{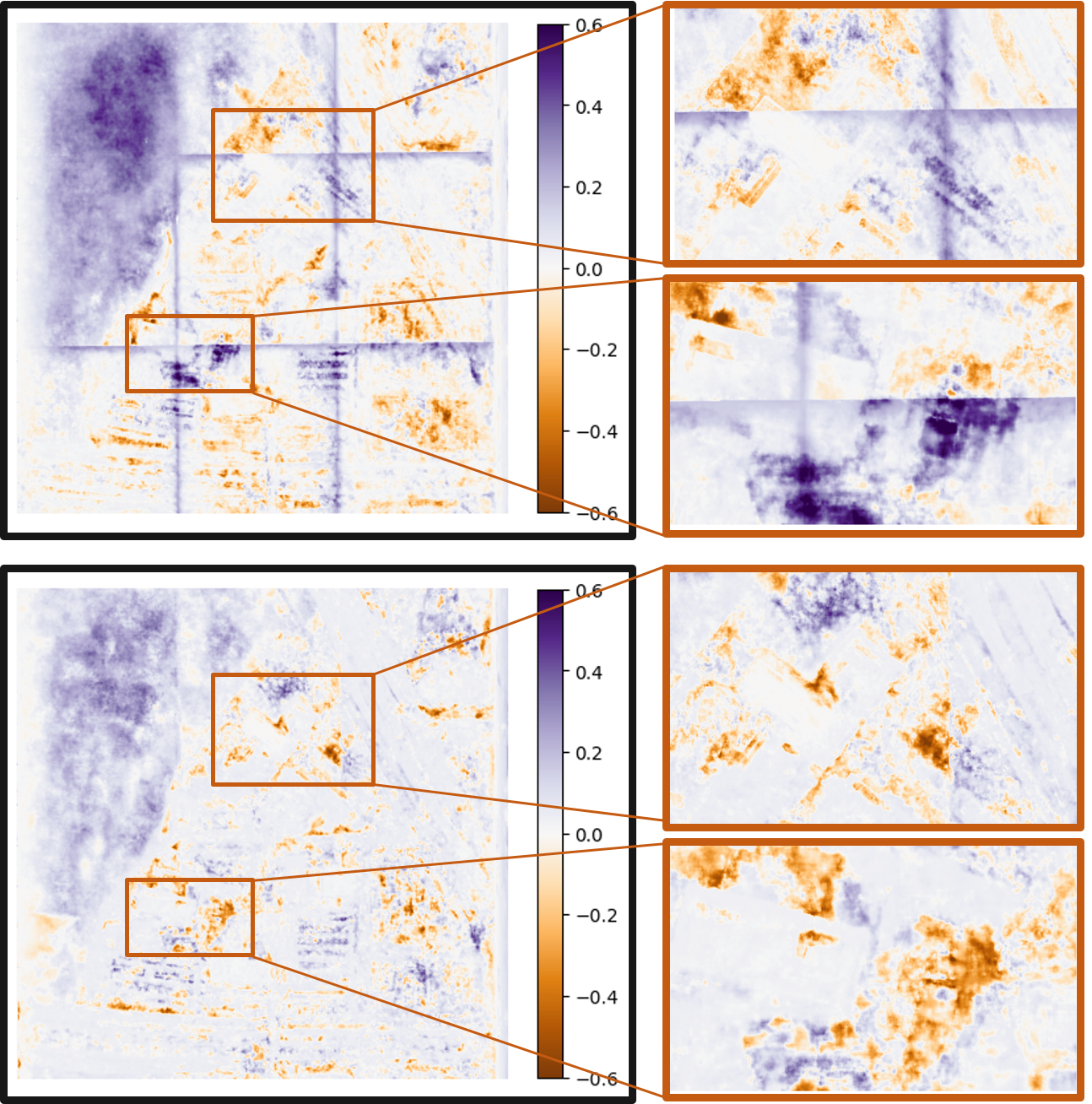}
\end{center}
   \caption{Differences between depth maps considering the reference algorithm. The \textbf{unscaled} case (\textbf{Top}), where each NeRFs is learned separately, hallucinates false matter along the 3D tile edges. For \textbf{Snake-NeRF} (\textbf{Bottom}), the differences are concentrated in areas that are ambiguous in the dataset (water, shadows, transient objects).}
\label{fig:diff_naif_9_22}
\end{figure}

The final row demonstrates our proposed scaling method. As shown in Table~\ref{tab:expe_metrics}, the NVS results closely resemble the unscaled version, achieving similar PSNR SSIM. The depth map contains no visible error patterns along the edges of the boundaries, indicating that our method effectively learns the continuity between neighboring domains of our NeRFs models. In fact, we observe that the differences between Snake-NeRF and the unscaled version (Figure~\ref{fig:diff_naif_9_22}, bottom) are concentrated on the areas where the NeRF is uncertain~\cite{goli2024bayes}. Given the simplistic NeRF model employed in this experiment, elements such as shadows, transient objects, and particularly water (visible in the top left of the \textbf{Middle} and \textbf{Bottom} depth maps) present inherent modeling challenges. These difficulties lead to instability across various model initializations. This explains the difference in Table~\ref{tab:expe_metrics}, between the \textbf{Reference case} and our \textbf{Scaled} method. 

% -------------------------------------------------------------------------

\section{Conclusion} \label{sec:conclusion}

% chaque rayons, quelque soit la structure de la grid, doit etre appris avec les nerfs qu'il intersecte. 

In conclusion, our research introduces an innovative solution to the challenge of scalability in 3D reconstruction using NeRF methods with high-resolution satellite imagery. Our approach is versatile and can be applied to scale various NeRF solutions proposed for satellite imagery, including Sat-NeRF~\cite{mari2022satnerf}, EO-NeRF~\cite{mari2023multi}, Season-NeRF~\cite{gableman2023incorporating}, and SUNDIAL~\cite{behari2024sundial}.

Our experiments show that scaling to large satellite images is possible with a NeRF framework, as long as: %under 2 conditions: 
\begin{itemize}
    \item Each ray is trained with all intersecting NeRFs in memory
    \item Each NeRF is continuously trained with sufficient information density across its entire input domain $(X, Y, Z)$
\end{itemize}

\section{Limitations and future work}

To validate our scaling algorithm in a controlled scenario, we utilized an open dataset that covers relatively small areas~\cite{lesaux2019data}. In fact, there are no publicly available datasets of multi-view satellite imagery over large areas on which we can release reproducible results, preventing further comparisons with multiple sets of data. Our hope is that the open-source release of our code will enable private owners of such datasets to benefit from our scaling methodology.

Additionally, our method does not incorporate the latest advancements in NeRFs for 3D Earth Observation, such as explicit modeling of shadows, transient objects, and albedo (or similar inverse rendering algorithms such as Gaussian Splatting~\cite{kerbl20233d}). Future work will explore the integration of these improvements within the context of scaling up. 

% Additionally, our method is based on uniform tiling without prior assumptions about the scene. We posit that semantics-aware tiling strategies could enhance the performance and quality of scene reconstruction. 

% Additionally, our method considers a constant min and max altitude on each tile. This might be very inefficient in case the scene contains very high objects, in which case we should use different altitudes per tile. This would require a first coarse reconstruction to estimate altitude bounds per tile.

%an initial coarse learning phase could provide valuable insights to adjust the altitude within different tiles and the ground coverage area per NeRF.
% For instance, smaller tiles could be used for city neighborhoods, while larger tiles for more homogeneous areas such as fields of cultivation. 

Finally, while we address the case for a single GPU, a multi-GPU extension could be beneficial to further speed-up training times on large areas. Across different altitudes, our cropping method derived from the RPC projection function can also be adapted to aerial applications.

% We believe Snake-NeRF also represents a step towards the fusion of satellite and other sources of imagery within a shared NeRF representation.

%utilizing a matrix camera model, which is pertinent to upcoming satellite missions, such as those involving matrix sensors like CO3D~\cite{10640872}. 

\newpage
\section{Acknowledgements}

This work was performed using HPC resources from CNES Computing Center (DOI 10.24400/263303/CNES\_C3). The elevations figures was computed using \href{https://xdem.readthedocs.io/en/stable/index.html}{xDEM} package (DOI 10.5281/zenodo.4809697). The authors would like to thank the Johns Hopkins University Applied Physics Laboratory and IARPA for providing the data used in this study, and the IEEE GRSS Image Analysis and Data Fusion Technical Committee for organizing the Data Fusion Contest. 

{\small
\bibliographystyle{ieee_fullname}
\bibliography{egbib}
}

\end{document}